\documentclass{article}
\usepackage{ijcai25}

\usepackage{times}
\usepackage{soul}
\usepackage{url}
\usepackage[hidelinks]{hyperref}
\usepackage[utf8]{inputenc}
\usepackage[small]{caption}
\usepackage{graphicx}
\usepackage{amsmath}
\usepackage{amsthm}
\usepackage{booktabs}
\usepackage{algorithm}
\usepackage{algorithmic}
\usepackage[switch]{lineno}

\usepackage{algorithm}
\usepackage{algorithmic}

\usepackage{amssymb}
\usepackage{amsmath}
\usepackage{multirow}
\usepackage[table,xcdraw]{xcolor}
\usepackage{xcolor}
\usepackage{colortbl}
\usepackage{tikz}
\usepackage{array}
\usepackage{booktabs}
\usepackage{makecell}
\usepackage{newfloat}
\usepackage{listings}

\title{T-T: Table Transformer for Tagging-based Aspect Sentiment Triplet Extraction}

\author{
    Anonymous submission
}

\author{
Kun Peng$^{1,2}$
\and
Chaodong Tong$^1$\and
Cong Cao$^1$\and
Hao Peng$^3$\and
Qian Li$^4$\and
Guanlin Wu$^5$\and\\
Lei Jiang$^1$\and
Yanbing Liu$^{1,2}$\And
Philip S. Yu$^6$
\\
\affiliations
$^1$Institute of Information Engineering, Chinese Academy of Sciences\\
$^2$School of Cyber Security, University of Chinese Academy of Sciences\\
$^3$School of Cyber Science and Technology, Beihang University \\
$^4$School of Computer Science, Beijing University of Posts and Telecommunications\\
$^5$College of Systems Engineering, National University of Defense Technology \\ 
$^6$Department of Computer Science, University of Illinois at Chicago\\
\emails
\{pengkun, tongchaodong\}@iie.ac.cn}

\begin{document}

\maketitle

\begin{abstract}
  Aspect sentiment triplet extraction (ASTE) aims to extract triplets composed of aspect terms, opinion terms, and sentiment polarities from given sentences. The table tagging method is a popular approach to addressing this task, which encodes a sentence into a 2-dimensional table, allowing for the tagging of relations between any two words. Previous efforts have focused on designing various downstream relation learning modules to better capture interactions between tokens in the table, revealing that a stronger capability to capture relations can lead to greater improvements in the model. Motivated by this, we attempt to directly utilize transformer layers as downstream relation learning modules. Due to the powerful semantic modeling capability of transformers, it is foreseeable that this will lead to excellent improvement. However, owing to the quadratic relation between the length of the table and the length of the input sentence sequence, using transformers directly faces two challenges: overly long table sequences and unfair local attention interaction. To address these challenges, we propose a novel Table-Transformer (T-T) for the tagging-based ASTE method. Specifically, we introduce a stripe attention mechanism with a loop-shift strategy to tackle these challenges. The former modifies the global attention mechanism to only attend to a 2-dimensional local attention window, while the latter facilitates interaction between different attention windows. Extensive and comprehensive experiments demonstrate that the T-T, as a downstream relation learning module, achieves state-of-the-art performance with lower computational costs.
\end{abstract}

\section{Introduction}\label{sec:introduction}

\begin{figure}[t]
    \centering
    \hspace{-0.3cm}\includegraphics[width=0.49\textwidth]{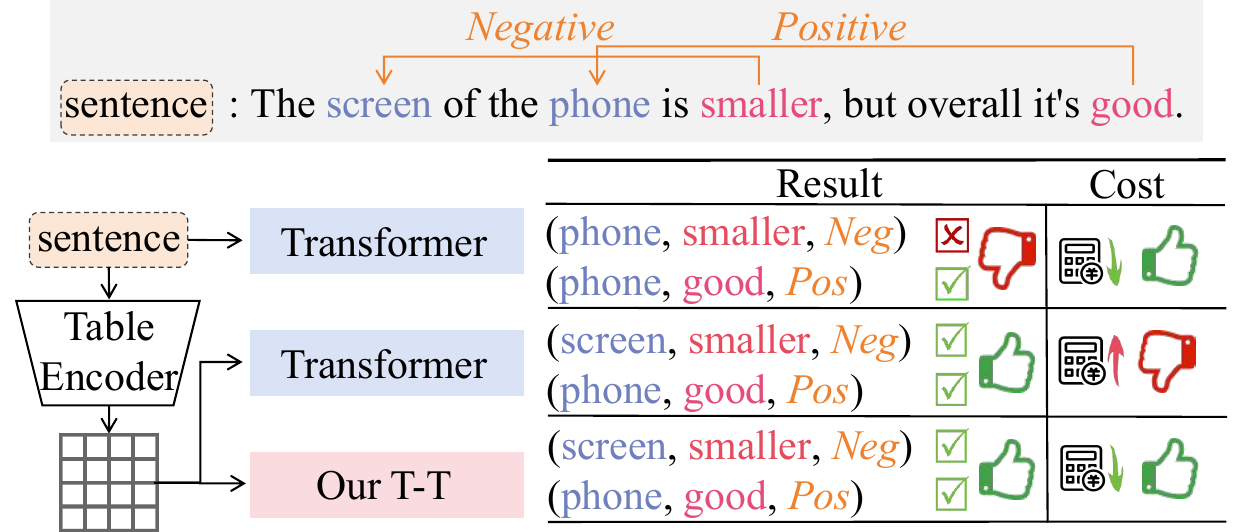}
    \vspace{-0.1cm}
    \caption{A toy example of three different ASTE methods. Our T-T model achieves commendable results in both performance and cost.}
    \label{fig:0}
\end{figure}

Aspect sentiment triplet extraction (ASTE) remains a crucial research direction in the era of large language models (LLMs) \cite{Wang2023IsCA,zhang-etal-2024-sentiment}, widely used for fine-grained opinion mining from user reviews, social media news, and other types of text \cite{zhang2022survey}.
ASTE task aims to extract corresponding aspect terms, opinion terms, and sentiment polarities from a given sentence. 
The sentiment polarity here is classified into three categories: \{Positive, Neutral, Negative\}. 
For example, as shown above in Figure \ref{fig:0}, given the sentence ``\textit{The screen of the phone is smaller, but overall it's good.}'', it has two sentiment triplets: (\textit{screen}, \textit{smaller}, Negative) and (\textit{phone}, \textit{good}, Positive), where \textit{screen} and \textit{phone} are the aspect terms, \textit{smaller} and \textit{good} are the corresponding opinion terms, Negative and Positive are the corresponding sentiment polarities.

Recently, a variety of techniques have been proposed for addressing the ASTE task, including span-based methods \cite{xu-etal-2021-learning,chen-etal-2021-semantic,li-etal-2023-dual-channel,zhao2024dual}, generative methods \cite{gao-etal-2022-lego,gou-etal-2023-mvp,xianlong-etal-2023-tagging,mukherjee-etal-2023-contraste,DBLP:journals/cogcom/ZouZWT24}, and tagging methods, among which the tagging method is particularly attractive.
Sequence tagging methods \cite{DBLP:conf/aaai/PengXBHLS20,xu-etal-2020-position,xianlong-etal-2023-tagging} use the BIESO scheme\footnote{BIESO means ``begin, inside, end, single, other'', respectively.} to tag aspect and opinion terms in sentences, but these methods fail to fully capture the relations between individual words.
Another more competitive tagging approach is the table tagging methods \cite{wu-etal-2020-grid,chen-etal-2022-enhanced,zhang-etal-2022-boundary,STAGE-2023,sun-etal-2024-minicongts}.
Given a sentence, it begins with encoding using a transformer-based pre-trained language model. 
Then, the obtained representations are encoded into a relation table.
In this relation table, the horizontal axis represents aspect terms, the vertical axis represents opinion terms, and each cell (e.g. \(cell_{ij}\)) in the table denotes the relation between the $i$-th and $j$-th words in the sentence. 
With this approach, sentiment triplets can be easily annotated within the table, and the relations between any two words can be fully captured.

One of the keys to the table tagging methods lies in designing different downstream relation encoding modules to help the model learn stronger relations representations. 
Previous methods \cite{wu-etal-2020-grid,jing-etal-2021-seeking,chen-etal-2022-enhanced,zhang-etal-2022-boundary,Liang2022STAGEST,peng2024prompt,peng2024table} have designed various downstream relation encoding modules, such as using Multi-Layer Perception (MLP) \cite{jing-etal-2021-seeking}, Graph Convolutional Network (GCN) \cite{chen-etal-2022-enhanced}, and Convolutional Neural Network (CNN) \cite{zhang-etal-2022-boundary}. 
These works demonstrate that a stronger relation encoding module can yield greater overall benefits for the model. 
This motivates us to seek even more powerful relation encoding modules.

In this work, we attempt to directly utilize transformer layers \cite{vaswani2017attention} as our downstream relation encoding module.
Ideally, the use of self-attention mechanisms in transformers enables the model to capture dependencies across the entire table sequences. 
This capability holds the promise of significantly enhancing the model's performance.
However, in reality, two formidable challenges impede the realization of this technique.
\textbf{Challenge 1: Overly long table sequences.} It is known that, if assuming the input sentence sequence length is $n$, the time and space complexity of the transformer layer in the multi-head self-attention module is $O(n^2)$. 
However, when dealing with table sequences, where the length of the table itself is $n^2$, the computational complexity of the attention mechanism in the transformer layer escalates to $O(n^4)$. This is unacceptable with limited computational resources.
\textbf{Challenge 2: Unfair local attention interaction.} Long sequences encourage us to use local attention mechanisms, but in reality, the correlation between tokens is not confined to local areas. When each token only focuses on its immediate periphery, capturing information from interactions with distant tokens becomes challenging.

To address the aforementioned challenges, we propose a novel Table-Transformer (T-T) for ASTE's table tagging method.
As shown in Figure \ref{fig:0}, when directly utilizing transformer-based sequence labeling models to process sentences, although cost-effective, their performance suffers due to their inability to fully capture word relations.
However, when encoding sentences into table sequences first and then using transformer layers as the relation encoder can yield good results, but the costs are intolerable due to the challenge of overly long sequences.  
Our approach, tailored for table sequence inputs, introduces unique enhancements to the original transformer layers, enabling the model to achieve strong performance while maintaining relatively low computational costs.
Specifically, to address the \textbf{overly long table sequences} challenge, we propose a stripe attention mechanism. It enhances the original global self-attention mechanism by restricting it to operate only within a fixed-size window range (assumed to be a constant $k$). This modification reduces the original time-space complexity from $O(n^4)$ to $O(k^2n^2)$.
For the \textbf{unfair local attention interaction} challenge, we devise a novel loop-shift strategy that effectively facilitates information interaction between different windows.

Our contributions can be summarized as follows:

1) We observe the two-dimensional nature of table sequences and based on this insight, we have designed an enhanced transformer layer called T-T. It can effectively capture local information within table sequences without relying on external knowledge or task-specific designs.

2) We introduce two novel techniques aimed at effectively addressing the two challenges encountered by the original transformer when handling table sequences.

3) Experimental results show that our approach achieves state-of-the-art performance within acceptable costs.

\section{Related Work}\label{sec:relatedwork}
\subsubsection{Aspect Sentiment Triplet Extraction.}
To meet the demand for a more detailed exploration of the opinion contained within the text, \cite{DBLP:conf/aaai/PengXBHLS20} first proposed and addressed the ASTE task in a pipeline manner.
Subsequently, more diversified techniques have been proposed.
MRC methods \cite{DBLP:conf/aaai/MaoSYC21,zhai-etal-2022-com,DBLP:journals/cogcom/ZouZWT24} treated the ASTE task as a form of machine reading comprehension task.
Generative methods \cite{zhang-etal-2021-towards-generative,gao-etal-2022-lego,gou-etal-2023-mvp,mukherjee-etal-2023-contraste} treated the ASTE task as an index generation task.
Span-based methods \cite{xu-etal-2021-learning,li-etal-2023-dual-channel,zhao2024dual} extracted all possible spans and considered the interplay of information at the span level. 

Table tagging, as another vibrant research direction, was initially proposed by GTS \cite{wu-etal-2020-grid} to annotate sentiment triplets in a 2D table. 
Subsequent extensive work, despite variations in the tagging schema, primarily focused on how to capture sufficient relation information for the table.
EMC-GCN \cite{chen-etal-2022-enhanced} utilized GCN to incorporate rich syntactic information. 
BDTF \cite{zhang-etal-2022-boundary} employed CNN to fully grasp word boundaries. 
STAGE \cite{Liang2022STAGEST} and SimSTAR \cite{10.1145/3539618.3592060} integrated the learning of span-level information into the relation encoder.
MiniConGTS \cite{sun-etal-2024-minicongts} proposed a new table tagging scheme based on contrastive learning.
In conclusion, the design of a more powerful downstream relation encoder stands as a pivotal focus in the table tagging methods.

\begin{figure*}[t]
    \centering
    \includegraphics[width=0.99\textwidth]{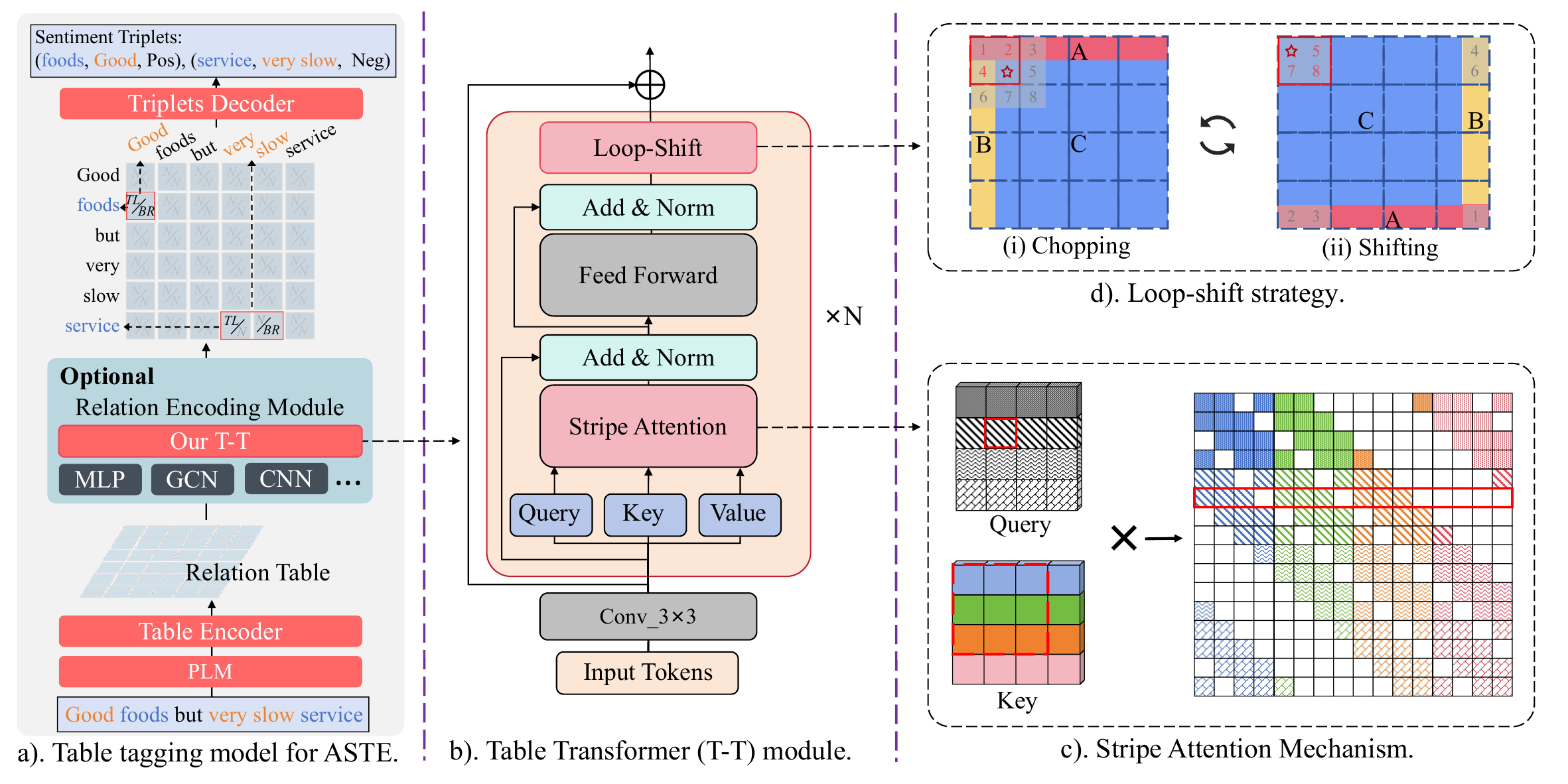}
    \vspace{-0.2cm}
    \caption{a) shows the architecture of our table tagging model for ASTE. One of the core components is the configurable relation encoding module. \textit{TL} and \textit{BR} donate the top-left and bottom-right vertex cells of the sentiment region, respectively. b) shows the architecture of our proposed T-T module. In sub-figure c), the left half shows the query matrix and key matrix divided into $4^2$ blocks, while the right half represents the attention map, where the colored blocks indicate the dot product computations. In sub-figure d), the final outputs of different layers shift between state (i) and state (ii). For a boundary token (marked with a red star), it attends to tokens 1, 2, and 4 in the output state (i), and to tokens 5, 7, and 8 in the output state (ii).}
    \label{fig:1}
\end{figure*}

\subsubsection{Efficient Transformer.}
The transformer model architecture \cite{vaswani2017attention} has become an indispensable tool in modern deep learning research due to its effectiveness in the field of NLP \cite{tay2022efficient}. 
However, the self-attention mechanism and stacked design logic of the transformer layers result in high computational resource requirements.
A large amount of previous work has focused on addressing this issue.
Some efforts attempted to reduce computational costs by modifying the scope of self-attention calculations.
These strategies include sliding windows \cite{2020Longformer,DBLP:conf/nips/ZaheerGDAAOPRWY20} or different attention mechanisms at various layers \cite{zhang-etal-2023-efficient-long}.
Other efforts have aimed at optimizing the computational efficiency of the self-attention matrix, such as applying a low-rank matrix \cite{wang2020linformer}, a kernel function \cite{katharopoulos2020transformers}, or a focused attention module \cite{han2023flatten}.
\section{Preliminary}\label{sec:preliminary}

\subsubsection{Task Definition.}
Given a sentence $X=\{x_1, x_2, ..., x_n\}$ of length $n$, the aim of ASTE is to extract all sentiment triplets $T=\{(a_1,o_1,s_1),$ $(a_2, o_2, s_2), ..., (a_{|T|}, o_{|T|}, s_{|T|})\}$ from $X$. Here, $a$ and $o$ respectively represent the aspect term and opinion term, which are spans within $X$, while $s$ denotes the corresponding sentiment polarity in \{\textit{Positive, Neutral, Negative}\}.

\subsubsection{Tagging Scheme.}
We choose the boundary-based tagging scheme \cite{zhang-etal-2022-boundary,ning-etal-2023-od} in our work.
For sentence $X$, it is mapped onto a two-dimensional table of size $n \times n$, where the vertical and horizontal axes represent aspect terms and opinion terms, respectively. 
For an sentiment triplet $(a_i, o_i, s_i)$ in $X$, assuming that $a_i$ and $o_i$ are located at positions $[x, y]$ and $[m, n]$ ($x \leq y, m \leq n$).
They can form a rectangular region in the table, with coordinates $(x, m)$ as the top-left corner and $(y, n)$ as the bottom-right corner.
We label the cells on the boundary of this region as \textit{TL} (top-left vertex) and \textit{BR} (bottom-right vertex), while all other cells are tagged as \textit{None}. 
When the sentiment elements are single words, the \textit{TL} and \textit{BR} positions coincide.
Subsequently, we label the sentiment polarity of this rectangular region according to $s_i$.

\subsubsection{Generalised Attention Mechanism.} 
For an input sequence $C = \{c_1, c_2, ..., c_n\} \in \mathbb{R}^{n \times d}$, the generalised attention mechanism of Transformer \cite{vaswani2017attention} treats the attention calculation over the entire sequence as a directed graph $G = \{C, A\}$, where the adjacency matrix $A \in \mathbb{R}^{n \times n}$ is used to describe directed edges. For any $a_{ij} \in A$ ($a_{ij}=1$ or $0$), if $a_{ij}=1$, it indicates that the query $c_i$ attends to the key $c_j$.
Let $N(c_i)$ denote all neighbors of $c_i$ (including itself), the attention calculation for $c_i$ is defined as:
\begin{equation}
ATTN(c_i) = \sigma (Q(c_i)K(N(c_i))^T)V(N(c_i)),
\label{attn}
\end{equation}
where $Q(\cdot), K(\cdot) : \mathbb{R}^{d} \rightarrow \mathbb{R}^{m}$ and $V(\cdot) : \mathbb{R}^{d} \rightarrow \mathbb{R}^{d}$ are the query, key, and value functions, respectively. $\sigma$ is a softmax function.

\section{Method}\label{sec:method}
Our framework is shown in Figure \ref{fig:1}, with the relation encoding module configured as our T-T\footnote{Codes are available at https://github.com/KunPunCN/T-T}.

\subsection{Table Encoding}
As illustrated in Figure \ref{fig:1} a), sentence $X$ is first inputted into a pre-trained language model (PLM) for encoding, yielding the last hidden layer's representation $H=\{h_1, h_2, ..., h_n\}$. 
Subsequently, we extract the features of aspects and opinions from $H \in \mathbb{R}^{n \times d}$ using two separate linear layers:
\begin{equation}
h^a_i = Linear_a(h_i), \quad h^o_j = Linear_o(h_j),
\end{equation}
where $h^a_i \in \mathbb{R}^d$ and $h^o_j \in \mathbb{R}^d$ denote the aspect and opinion representation, respectively.
To ensure that the table captures sufficient relation information, which is beneficial for downstream tasks, we followed the previous works \cite{chen-etal-2022-enhanced}, employing Biaffine Attention \cite{dozat2017deep} to capture the relations distribution between words.
This process is formalized as:
\begin{equation}
r_{ij} = \boldsymbol{W_1}({h^a}_i \oplus (h^o)_j) \oplus {{h^a}_i}^T\boldsymbol{W_2}{h^o}_j \oplus Pooling(h_{[i,j]}),
\end{equation}
where $\oplus$ denotes the concatenation operation.
The first two terms represent the biaffine attention process, where $\boldsymbol{W_1} \in \mathbb{R}^{d \times 2d}$ and $\boldsymbol{W_2} \in \mathbb{R}^{d \times \sqrt{d} \times d}$ are learnable parameters.
The last term $Pooling(h_{[i,j]})$ represents the maxpooling operation for the sentence from token $h_i$ to $h_j$, which aims to extract the span-level information.
$r_{ij} \in \mathbb{R}^{2d + \sqrt{d}}$ is the table representation. 
Finally, a dense linear layer $Linear_d$: $\mathbb{R}^{2d + \sqrt{d}} \rightarrow \mathbb{R}^{d'}$ is used to compress $r_{ij}$'s dimension to $d'$: $r'_{ij} = Linear_d(r_{ij})$.
Therefore, the final representation of the relation table is denoted as $R \in \mathbb{R}^{n \times n \times d'}$.

\subsection{Relation Encoding with T-T module}
The initialized relation table needs additional refinement of the relations distribution through the proposed T-T module.
As depicted in Figure \ref{fig:1} b), we first use a simple convolutional layer with a kernel size of 3 to capture spatial correlations, and then feed the obtained representation $R^0 \in \mathbb{R}^{n \times n \times d'}$ into the T-T module.
The T-T improves upon the original transformer architecture by replacing the full attention mechanism with an enhanced stripe attention mechanism and adding a novel loop-shift strategy at the end of each layer.

\subsubsection{Stripe Attention Mechanism.}
Here we propose a novel stripe attention mechanism to address the overly long table sequence challenge.
As described in the Preliminary section, the generalised attention mechanism treats the attention calculation as a directed graph $G = \{C, A\}$.
In the original transformer layer, a full attention mechanism is used, which means $A$ is the full ones matrix and $N(c_i) = C$ in Formula (1).
When the input is a sequence $C \in \mathbb{R}^{n \times d'}$, each token needs to interact with all other tokens and store all computation results, leading to a spatiotemporal complexity of $O(n^2)$.
The derivation reveals that the spatiotemporal complexity of the full attention layer scales quadratically with the length of the input sequence.
When our input is the flattened table sequence $R^0 \in \mathbb{R}^{n \times n \times d'}$, the overall spatiotemporal complexity becomes $O(n^4)$. 
This is intolerable for limited computational resources.

To maintain good performance while reducing the spatiotemporal complexity, the proposed stripe attention mechanism partitions the table sequence $R^0$ into multiple smaller blocks, treating them as the smallest computational units.
Assuming each block has a width of $b$ and a shape of $\mathbb{R}^{b \times b \times d'}$ (where $n$ is padded beforehand to a multiple of $b$), this results in a total of $l^2$ (we define $l=\frac{n}{b}$) blocks: $O = \{o_1, o_2, ..., o_{l^2}\} \in \mathbb{R}^{l^2 \times d'}$.
It is worth noting that we apply a cyclic shift operation on this sequence, which means that for a block $o_i$ in this sequence, its left and right neighbors in the attention map are $o_{i\pm1} = o_{(i\pm1 + l^2)\%l^2}$, and its upper and lower neighbors are $o_{i\pm l} = o_{(i\pm l + l^2)\%l^2}$, where \% is used for the remainder calculation.
Similar to Formula \ref{attn}, for each block $o_{i}$, it's attention calculation is defined as:
\begin{equation}
ATTN(o_{i}) = \sigma (Q(o_{i})K(N(o_{i}))^T)V(N(o_{i})).
\label{attn_b}
\end{equation}
Per Formula \ref{attn_b}, the query matrix and the key matrix require a dot product operation to generate the attention map.

As shown in the left half of Figure \ref{fig:1} c), the query and key matrices are also partitioned into multiple blocks. 
According to Formula \ref{attn_b}, a dot product operation is needed for the query matrix and the key matrix to generate the attention map. 
In the full attention mechanism, each query block needs to undergo a dot product operation with all key blocks, resulting in a complexity of $O(n^4)$.
To enhance attention performance while maintaining lower complexity, in our stripe attention mechanism, each query block only selectively attends to its neighboring blocks in the key matrix.
As indicated by the highlighted red boundary box in Figure \ref{fig:1} c), for a query block, we define its neighbors in the 2D table space as all key blocks within a square area centered around it.
Assuming a window width of $w (w \leq l)$ centered around each block for attention, then the neighbors of block $o_{i}$ can be defined as:
\begin{equation}
N(o_{i}) = \begin{Bmatrix}
    o_{i-\lfloor \frac{w}{2} \rfloor l -\lfloor \frac{w}{2} \rfloor}, & ... & o_{i-\lfloor \frac{w}{2} \rfloor l+\lfloor \frac{w}{2} \rfloor}, \\
    o_{i-(\lfloor \frac{w}{2} \rfloor-1) l -\lfloor \frac{w}{2} \rfloor}, & ... & o_{i-(\lfloor \frac{w}{2} \rfloor-1) l+\lfloor \frac{w}{2} \rfloor}, \\
    \vdots&\vdots&\vdots \\
    o_{i+\lfloor \frac{w}{2} \rfloor l -\lfloor \frac{w}{2} \rfloor}, & ... & o_{i+\lfloor \frac{w}{2} \rfloor l +\lfloor \frac{w}{2} \rfloor} \\
\end{Bmatrix},
\label{nnn}
\end{equation}
where $w$ is odd. 
The number of $N(x_i)$ is  $|N(x_i)| = w^2$, which means that for each block (total of $n^2$ tokens), it attends to all blocks within a square area of $w^2$, encompassing a total of $w^2 \times b^2$ tokens. 
The overall complexity is $O(n^2 \times w^2b^2) = O(w^2b^2n^2)$.
In addition, when $w=l$, $N(x_{i})$ will include all blocks, and stripe attention will degenerate into full attention.

An example of the stripe attention map is shown in the right half of Figure \ref{fig:1} (c), where $n=4b$ and $w=3$. 
Compared to local attention, stripe attention allows tokens to attend to a broader range of relevant positions in the 2D table space, thereby enhancing learning capability, while only marginally increasing complexity by a constant factor.

\subsubsection{Loop-shift Strategy.}

Here we propose an innovative loop-shift strategy to address the unfair local attention interaction challenge. 
While the stripe attention mechanism achieves a balance between attention scope and computational load, partitioning the table sequence into fixed block widths prohibits interaction between different blocks. 
This presents an unfair scenario for tokens situated at block edges, as their crucial neighbors may reside in different blocks.
Therefore, inspired by \cite{DBLP:conf/iccv/LiuL00W0LG21}, we introduce a novel loop-shift operation to the table sequence between different T-T layers.
For the output of any given layer, which has a shape of $n \times n$, before feeding it into the next layer, we first transform the positional distribution of the table tokens.
As illustrated in Figure \ref{fig:1} d), we perform a two-step shift operation on the table with a width of $\frac{b}{2}$. 
Firstly, we chop the tokens from the top of the table, forming a shape of $\frac{b}{2} \times n$, and move them to the bottom of the table. 
Secondly, we chop the tokens on the left side of the table, forming a shape of $n \times \frac{b}{2}$, and move them to the right side of the table. 
This operation effectively cyclically shifts the entire table to the upper-left direction by a distance of $\frac{b}{2}$.
Similarly, for the next layer’s output, we need to cyclically shift the entire table in the opposite direction, by a distance of $\frac{b}{2}$ towards the bottom-right direction, to restore the table's distribution.
Therefore, the shift operation occurs in pairs, implying that the number of T-T layers in our model is even.

After passing through $N$ layers of the T-T module, we obtain the last layer's representation $R^N \in \mathbb{R}^{n \times n \times d'}$. 
Then a weighted residual connection is used to derive the final representation: $R = \boldsymbol{W_3} R^N + (1-\boldsymbol{W_3}) R^0$.

\subsection{Triplet Decoding}
During the decoding stage, we employ two separate full linear layers to predict the coordinates of all \textit{TL} and \textit{BR}, respectively. 
The predicted results for each position are $p_{ij}^{TL}$ and $p_{ij}^{BR}$, respectively.
Given the labels $y_{ij}^{TL}$ and $y_{ij}^{BR}$, the corresponding loss is:
\begin{equation}
\mathcal{L}_1 = \sum_{i=1}^n\sum_{j=1}^nL_{CE}(p_{ij}^{TL}, y_{ij}^{TL}) + L_{CE}(p_{ij}^{BR}, y_{ij}^{BR}),
\end{equation}
where $L_{CE}$ is the cross entropy loss function.
Subsequently, for any pair of \textit{TL} and \textit{BR}, they form a candidate rectangle when the \textit{TL} is positioned to the upper left or coincides with the \textit{BR}.
This rectangle is represented as:
\begin{equation}
r_{abcd} = r_{ab} \oplus r_{cd} \oplus pooling(r_{ab},...,r_{cd}),
\end{equation}
where $r_{ab}$ and $r_{cd}$ respectively represent the token at \textit{TL} and \textit{BR}, $r_{ab},...,r_{cd}$ denote all tokens within this rectangle.
Then, a sentiment classifier $Linear_s$: $\mathbb{R}^{3d} \rightarrow \mathbb{R}^{4}$ is utilized to predict the sentiment polarity of this region as one of \{ \textit{Pos, Neu, Neg, Invalid}\}: $p_{k}^{s} = Linear_s(r_{abcd})$.
Since each \textit{TL/BR} may form candidate rectangles with multiple \textit{BR/TL}, the label \textit{Invalid} is additionally used to assess the validity of the candidate region.
When the prediction $p_{k}^{s}$ is \textit{Invalid}, the region should be dropped.
When the prediction $p_{k}^{s}$ falls within \{ \textit{Pos, Neu, Neg}\}, the corresponding sentiment triplet can be extracted based on the \textit{TL}, \textit{BR} coordinates, and sentiment polarity of the region.
The loss of this process is:
\begin{equation}
\mathcal{L}_2 = \sum_{k=1}^mL_{CE}(p_{k}^{s}, y_{k}^{s}),
\end{equation}
where $m$ represents the number of candidate rectangles, $p_{k}^{s}$ and $y_{k}^{s}$ are the prediction and label, respectively, for the $k$-th candidate rectangles.
The overall loss is: $\mathcal{L} = \mathcal{L}_1 + \mathcal{L}_2$.

\section{Experiments}\label{sec:experiments}

\subsection{Experimental Settings}
\subsubsection{Datasets and Baselines.}
We conduct our experiments on four benchmark datasets \cite{DBLP:conf/aaai/PengXBHLS20,xu-etal-2020-position}, which are originally derived from the SemEval challenge \cite{pontiki-etal-2014-semeval,pontiki-etal-2015-semeval,pontiki-etal-2016-semeval}.
The detailed statistics are provided in Table \ref{tab:statistics}.
We categorize the comparison baselines into five types: seq tagging, MRC-based, generative, span-based, and table tagging.

\begin{table}[t]\small
\centering
\tabcolsep=4pt
\belowrulesep=0pt
\aboverulesep=0pt
\scalebox{1}{
\begin{tabular}{lcccccccccc}
\Xhline{0.9pt}
\multicolumn{2}{c}{\multirow{2}{*}{\textbf{Dataset}}} 
    & \multicolumn{2}{c}{\textbf{14res}} & \multicolumn{2}{c}{\textbf{14lap}} & \multicolumn{2}{c}{\textbf{15res}} & \multicolumn{2}{c}{\textbf{16res}} \\ 
    \cmidrule(lr){3-4}\cmidrule(lr){5-6}\cmidrule(lr){7-8}\cmidrule(lr){9-10}
    & & \#S        & \#T          & \#S         & \#T         & \#S        & \#T          & \#S        & \#T          \\ \hline
\multirow{3}{*}{}              
    & train            & 1,266       & 2,338               & 906        & 1460        & 605        & 1,013        & 857        & 1,394        \\
    & dev           & 310         & 577              & 219        & 346          & 148        & 249          & 210        & 339          \\
    & test            & 492         & 994              & 328        & 543        & 322        & 485          & 326        & 514          \\ \Xhline{0.9pt}
\end{tabular}
}
\caption{Statistics of datasets, where \#S and \#T represent the number of sentences and triplets, respectively.}
\label{tab:statistics}
\end{table}

\begin{table*}[t]\small
\centering
\belowrulesep=0pt
\aboverulesep=0pt
\tabcolsep=3.3pt
\begin{tabular}{ccccccccccccc|c}
\Xhline{0.9pt}
\multicolumn{1}{c}{\multirow{2}{*}{\textbf{Model}}}
    & \multicolumn{3}{c}{14res}      & \multicolumn{3}{c}{14lap}      & \multicolumn{3}{c}{15res}           & \multicolumn{3}{c|}{16res}  & \multirow{2}{*}{Avg-F1}   \\ 
    \cmidrule(lr){2-4}\cmidrule(lr){5-7}\cmidrule(lr){8-10}\cmidrule(lr){11-13}
\multicolumn{1}{l}{} & P.    & R.        & F1             & P.    & R.    & F1             & P.    & R.    & F1             & P.    & R.    & F1   &{}      \\ \hline
\multicolumn{1}{c}{\multirow{1}{*}{\textbf{Seq tagging}}}& \\
\multicolumn{1}{c}{Peng-two-stage \cite{DBLP:conf/aaai/PengXBHLS20}}                  & 43.24 & 63.66 & \cellcolor[HTML]{EFEFEF}51.46          & 37.38 & 50.38 & \cellcolor[HTML]{EFEFEF}42.87          & 48.07 & 57.51 & \cellcolor[HTML]{EFEFEF}52.32          & 46.96 & 64.24 & \cellcolor[HTML]{EFEFEF}54.21  &\cellcolor[HTML]{EFEFEF}50.22    \\ 
\multicolumn{1}{c}{TAGS$^\ddagger$ \cite{xianlong-etal-2023-tagging}}        &74.92 &73.81 &\cellcolor[HTML]{EFEFEF}74.36    &64.69 & 61.89 &\cellcolor[HTML]{EFEFEF}63.26   &69.55 &65.25 &\cellcolor[HTML]{EFEFEF}67.33 &75.40 &72.48 &\cellcolor[HTML]{EFEFEF}74.17  &\cellcolor[HTML]{EFEFEF}69.78   \\
\hline

\multicolumn{1}{c}{\multirow{1}{*}{\textbf{MRC-based}}}\\

\multicolumn{1}{c}{COM-MRC \cite{zhai-etal-2022-com}}                          & 75.46 & 68.91 & \cellcolor[HTML]{EFEFEF}72.01          & 62.35 & 58.16 & \cellcolor[HTML]{EFEFEF}60.17          & 68.35 & 61.24 & \cellcolor[HTML]{EFEFEF}64.53          & 71.55 & 71.59 & \cellcolor[HTML]{EFEFEF}71.57 &\cellcolor[HTML]{EFEFEF}67.07 \\

\multicolumn{1}{c}{Triple-MRC \cite{DBLP:journals/cogcom/ZouZWT24}}                          & / & / & \cellcolor[HTML]{EFEFEF}72.45          & / & / & \cellcolor[HTML]{EFEFEF}60.72          & / & / & \cellcolor[HTML]{EFEFEF}62.86          & / & / & \cellcolor[HTML]{EFEFEF}68.65 &\cellcolor[HTML]{EFEFEF}66.17

\\
\hline
\multicolumn{1}{c}{\multirow{1}{*}{\textbf{Generative}}}\\
\multicolumn{1}{c}{GPT3.5$^\ddagger$ \cite{ouyang2022training}}                        & 60.77 & 54.22 & \cellcolor[HTML]{EFEFEF}57.31        & 59.13 & 49.27 & \cellcolor[HTML]{EFEFEF}53.76      & 58.24 & 48.23 & \cellcolor[HTML]{EFEFEF}52.77   & 59.33 & 56.02 & \cellcolor[HTML]{EFEFEF}57.60 &\cellcolor[HTML]{EFEFEF}55.36 \\
\multicolumn{1}{c}{LEGO \cite{gao-etal-2022-lego}}     & / & / & \cellcolor[HTML]{EFEFEF}73.7          & / & / & \cellcolor[HTML]{EFEFEF}62.2          & / & / & \cellcolor[HTML]{EFEFEF}64.4          & / & / & \cellcolor[HTML]{EFEFEF}69.9  &\cellcolor[HTML]{EFEFEF}67.55 \\
%
\multicolumn{1}{c}{MvP \cite{gou-etal-2023-mvp}}        &/ &/ &\cellcolor[HTML]{EFEFEF}74.05    &/ &/ &\cellcolor[HTML]{EFEFEF}63.33   &/ &/ &\cellcolor[HTML]{EFEFEF}65.89 &/ &/ &\cellcolor[HTML]{EFEFEF}73.48  &\cellcolor[HTML]{EFEFEF}69.19  \\  
\multicolumn{1}{c}{CONTRASTE \cite{mukherjee-etal-2023-contraste}}        &73.6 &74.4 &\cellcolor[HTML]{EFEFEF}74.0    &64.2 &\textbf{61.7} &\cellcolor[HTML]{EFEFEF}62.9   &65.3 &\textbf{66.7} &\cellcolor[HTML]{EFEFEF}66.1 &72.2 &76.3 &\cellcolor[HTML]{EFEFEF}74.2  &\cellcolor[HTML]{EFEFEF}69.3 \\   
\hline
\multicolumn{1}{c}{\multirow{1}{*}{\textbf{Span-based}}}\\
\multicolumn{1}{c}{Span-ASTE \cite{xu-etal-2021-learning}}                  & 72.89 & 70.89 & \cellcolor[HTML]{EFEFEF}71.85          & 63.44 & 55.84 & \cellcolor[HTML]{EFEFEF}59.38          & 62.18 & 64.45 & \cellcolor[HTML]{EFEFEF}63.27 & 69.45 & 71.17 & \cellcolor[HTML]{EFEFEF}70.26  &\cellcolor[HTML]{EFEFEF}66.19  \\ 
\multicolumn{1}{c}{D2E2S \cite{zhao2024dual}}         &75.92 &74.36 &\cellcolor[HTML]{EFEFEF}75.13    &67.38 &60.31 &\cellcolor[HTML]{EFEFEF}\underline{63.65}   &70.09 &62.11 &\cellcolor[HTML]{EFEFEF}65.86 &\textbf{77.97} &71.77 &\cellcolor[HTML]{EFEFEF}74.74 &\cellcolor[HTML]{EFEFEF}69.84 \\   
\hline

\multicolumn{1}{c}{\multirow{1}{*}{\textbf{Table tagging}}}\\
\multicolumn{1}{c}{GTS-BERT$^\dagger$ \cite{wu-etal-2020-grid}}                        & 68.09 & 69.54 & \cellcolor[HTML]{EFEFEF}68.81          & 59.40 & 51.94 & \cellcolor[HTML]{EFEFEF}55.42          & 59.28 & 57.93 & \cellcolor[HTML]{EFEFEF}58.60          & 68.32 & 66.86 & \cellcolor[HTML]{EFEFEF}67.58   &\cellcolor[HTML]{EFEFEF}62.60 \\%
\multicolumn{1}{c}{EMC-GCN \cite{chen-etal-2022-enhanced}}                          & 71.21 & 72.39 & \cellcolor[HTML]{EFEFEF}71.78          & 61.70 & 56.26 & \cellcolor[HTML]{EFEFEF}58.81          & 61.54 & 62.47 & \cellcolor[HTML]{EFEFEF}61.93          & 65.62 & 71.30 & \cellcolor[HTML]{EFEFEF}68.33   &\cellcolor[HTML]{EFEFEF}65.21  \\

\multicolumn{1}{c}{BDTF \cite{zhang-etal-2022-boundary}}                          &75.53 &73.24 &\cellcolor[HTML]{EFEFEF}74.35     &\underline{68.94} &55.97 &\cellcolor[HTML]{EFEFEF}61.74    &68.76 &63.71 &\cellcolor[HTML]{EFEFEF}66.12    &71.44 &73.13 &\cellcolor[HTML]{EFEFEF}72.27 &\cellcolor[HTML]{EFEFEF}68.62 \\
\multicolumn{1}{c}{STAGE-3D \cite{Liang2022STAGEST}}                          &\textbf{78.58} &69.58 &\cellcolor[HTML]{EFEFEF}73.76     &\textbf{71.98} &53.86 &\cellcolor[HTML]{EFEFEF}61.58    &\textbf{73.63} &57.90 &\cellcolor[HTML]{EFEFEF}64.79    &\underline{76.67} &70.12 &\cellcolor[HTML]{EFEFEF}73.24 &\cellcolor[HTML]{EFEFEF}68.34 \\ 
\multicolumn{1}{c}{SimSTAR \cite{10.1145/3539618.3592060}}                          &76.23 &71.63 &\cellcolor[HTML]{EFEFEF}73.86     &66.46 &58.23 &\cellcolor[HTML]{EFEFEF}62.07    &71.71 &59.59 &\cellcolor[HTML]{EFEFEF}65.09    &72.07 &74.12 &\cellcolor[HTML]{EFEFEF}73.06 &\cellcolor[HTML]{EFEFEF}68.52 \\ 
\multicolumn{1}{c}{MiniConGTS \cite{sun-etal-2024-minicongts}}        &76.10 &\textbf{75.08} &\cellcolor[HTML]{EFEFEF}\underline{75.59}    &66.82 & \underline{60.68} &\cellcolor[HTML]{EFEFEF}63.61   & 66.50 &63.86 &\cellcolor[HTML]{EFEFEF}65.15 & 75.52 &\underline{74.14} &\cellcolor[HTML]{EFEFEF}\underline{74.83} &\cellcolor[HTML]{EFEFEF}69.80 \\
\hline
\multicolumn{1}{c}{\textbf{T-T (Ours)}}       &\underline{77.06}  &\underline{74.56}  & \cellcolor[HTML]{EFEFEF}\textbf{75.79} &67.50 &60.25 & \cellcolor[HTML]{EFEFEF}\textbf{63.67} &\underline{72.15}  &\underline{65.40}  & \cellcolor[HTML]{EFEFEF}\textbf{68.61}  &75.05 &\textbf{74.63} & \cellcolor[HTML]{EFEFEF}\textbf{74.84} &\cellcolor[HTML]{EFEFEF}\textbf{70.73}\\ 
\multicolumn{1}{c}{\quad -w/ Random init.}       &76.24  &73.44  & \cellcolor[HTML]{EFEFEF}74.81 &66.61 &60.19 & \cellcolor[HTML]{EFEFEF}63.24 &71.17  &65.39  & \cellcolor[HTML]{EFEFEF}\underline{68.16}         &74.60  &73.19 & \cellcolor[HTML]{EFEFEF}73.89 &\cellcolor[HTML]{EFEFEF}\underline{70.03} \\ 
\Xhline{0.9pt}
\end{tabular}
\caption{Main results on 4 datasets. $\dagger$, $\ddagger$ denote that results are obtained from [Chen \textit{et al}., 2022] and conducted by us, other results are from the original papers. The best results are in \textbf{bold}, while the second best are \underline{underlined}.}
\label{tab:mainresult}
\end{table*}

\newcommand{\mycell}[3]{%
  \begin{tikzpicture}[baseline=(C.base)]
    \node[fill=#1, rectangle, inner sep=0pt, minimum width=#2, minimum height=3mm] (C) {\textcolor{black}{#3}};
  \end{tikzpicture}
}

\begin{table*}[t]
\centering
\begin{tabular}{l|ccccccc}
\Xhline{0.9pt}
  \multicolumn{1}{c}{\textbf{Ablation Settings}}    & \#Param.        & \#Training Cost & \#Interfere Cost & 14res F1 & 14lap F1 & 15res F1 & 16res F1 \\ \hline
Full Model            & 184.3M          & {5.93 /ms}        & {1.44 /ms}         & {75.79}    & {63.67}    & {68.61}    & {74.84}    \\
 \hline
-w/o Loop-shift       & 184.3M          &{5.53 /ms}        & {1.33 /ms}         & \mycell{gray!25}{1cm}{73.63}    & \mycell{gray!25}{1cm}{62.35}    & \mycell{gray!25}{1cm}{66.38}    & \mycell{gray!25}{1cm}{72.73}    \\
-w/o Stripe Attention (SA)  & 184.3M    &  \mycell{gray!25}{1.4cm}{19.14 /ms}     & \mycell{gray!25}{1.4cm}{4.40 /ms}        & {75.81}    & {63.89}    & {68.73}    & {74.80}    \\
-w/ Normal layers  & 184.3M    &  \mycell{gray!25}{1.4cm}{19.03 /ms}     & \mycell{gray!25}{1.4cm}{4.32 /ms}        & {75.74}    & {63.81}    & {68.52}    & {74.68}    \\
-w/o T-T Relation Encoder & 159.1M          & {4.74 /ms}       &  {1.12 /ms}        & \mycell{gray!25}{1cm}{71.42}    & \mycell{gray!25}{1cm}{60.28}    & \mycell{gray!25}{1cm}{64.67}   & \mycell{gray!25}{1cm}{70.51}    \\ 
\Xhline{0.9pt}
\end{tabular}
\caption{Ablation results on 4 datasets. ``-w/o'' means without. ``/ms'' donates the average computation time per sentence.}
\label{tab:ablation}
\end{table*}

\subsubsection{Implementation.}
The proposed model contains a table encoder and a relation encoder, with hidden state dimensions of $d=768$ and $d'=1024$, respectively.
We initialize the table encoder with the BERT-base-uncased \cite{devlin-etal-2019-bert} version.
As a relation encoder, our T-T module consists of two transformer layers and utilizes the parameters from the last two layers of the BERT-large version for initialization.
The block width $b$ is 7 for 14res, 15res, 16res and 5 for 14lap.
The window width $w$ is 3.
During training, we use the AdamW optimizer with an initial learning rate of 3e-5 for all layers.
The model is trained for 15 epochs on RTX 3090 GPUs, with a batch size of 4.
In each epoch, we evaluate the training model on the development set and save the best one.
We use the sentence-level F1 score as the evaluation metric, which means that a sentence is considered a true positive only when all triples within it are correctly extracted.
All the reported results are the average of five runs with different random seeds.

\subsection{Main Results}
As shown in Table \ref{tab:mainresult}, we have the following observations:
Our model surpassed the best baseline, D2E2S, with improvements of 0.66\%, 0.02\%, 2.75\%, and 0.1\% F1 scores on 14res, 14lap, 15res, and 16res, respectively, resulting in an overall average improvement of 0.89\% F1 score. 
While D2E2S introduced additional tools to provide syntactic dependency information and developed task-specific unique modules, our model solely utilized a streamlined architecture to effectively capture local information within 2D table sequences.

Our model significantly outperformed all table tagging methods. Compared to the best table-based model, MiniConGTS, our model surpassed it by 0.2\%, 0.07\%, 3.46\%, and 0.01\% F1 scores on 14res, 14lap, 15res, and 16res, respectively, under the F1 scores, with an overall average improvement of 0.93\%. 
This demonstrates that our T-T module exhibits stronger relation encoding capabilities compared to previous relation encoding modules.

Compared to initializing the T-T parameters using the last two layers of BERT, ``Random init." indicates random initialization of the T-T, which resulted in an average decrease of 0.70\% F1 score in overall model performance. 
We attribute this to the insufficient training data size to optimize the parameters of the two transformer layers, resulting in model underfitting. 
On the other hand, initializing with the last two layers of pre-trained BERT incorporates certain semantic knowledge, thereby improving the model's effectiveness.

\subsection{Ablation Study and Computational Cost}
We conduct an ablation experiment to validate the proposed module's improvements in model performance and costs.
As shown in Table \ref{tab:ablation}, since loop-shift and Stripe Attention (SA) are parameter-independent mechanisms, removing them will not alter the model's parameters. 

When the loop-shift strategy is removed, the training and inference costs remain largely unchanged. 
However, the performance decreased by 2.16\%, 1.32\%, 2.23\%, and 2.11\% F1 scores on the four datasets, demonstrating the crucial importance of information interaction across different blocks for enhancing the overall model performance.
When SA is removed, the model degrades to full attention while retaining the Loop-shift strategy, resulting in performance similar to that of directly using normal layers.
After removing SA, we observe a slight improvement in model performance, but there is a significant deterioration in both training and inference costs, which are 13.21/ms and 2.96/ms, respectively. 
This demonstrates that the Stripe Attention mechanism not only greatly improves attention computation efficiency but also effectively attends to relevant tokens.
When the T-T relation encoder module is removed, the model's performance significantly deteriorates, with respective drops of 4.37\%, 3.39\%, 3.94\%, and 4.33\%. 
This demonstrates the critical importance of designing a stronger relation encoding module.

\begin{table}[t]
\begin{tabular}{c|ccccc}
\Xhline{0.9pt}
\multicolumn{1}{c}{\textbf{Model}}   & 14res & 14lap & 15res & 16res & Avg-F1 \\ \hline
GTS$^\dagger$     & 74.63 & 66.46 & 67.52 & \multicolumn{1}{c|}{74.20} & 70.70   \\
EMC-GCN$^\dagger$ & 76.33 & 67.94 & 67.26 & \multicolumn{1}{c|}{74.15} & 71.42   \\
STAGE$^\dagger$   & 77.87 & 69.70 & 70.60 & \multicolumn{1}{c|}{\underline{79.98}} & 74.54   \\ 
MiniConGTS   & \underline{79.60} & \underline{73.23} & \underline{73.87} & \multicolumn{1}{c|}{76.29} & \underline{75.75}   \\ \hline
T-T     & \textbf{79.97} & \textbf{73.40} & \textbf{74.35} & \multicolumn{1}{c|}{\textbf{80.11}} & \textbf{76.96}   \\ \Xhline{0.9pt}
\end{tabular}
\caption{Results on the AOPE task. $\dagger$ denotes that results are derived from [Liang \textit{et al}., 2023a]. The best results are in \textbf{bold}, while the second best are \underline{underlined}.}
\label{tab:aope}
\end{table}

\subsection{Auxiliary Experiment on Subtask}
To further investigate the effectiveness of T-T, we conduct an auxiliary experiment on the Aspect Opinion Pair Extraction (AOPE) task, which aims to extract all aspect-opinion pairs from a sentence.
By modifying the category width of the sentiment classifier $Linear_s$ from 4 (\textit{\{Pos, Neu, Neg, invalid}\}) to 2 (\textit{\{Valid, invalid}\}), our model can directly address these tasks without any additional modifications.
As depicted in Table \ref{tab:aope}, we chose the table tagging methods for comparison to demonstrate the specific improvements of our model.
The results demonstrate that our T-T achieves comprehensive improvements across four datasets, with an average F1 score increase of 0.96\% compared to the best baseline, MiniConGTS. 
This further validates the important role of T-T in capturing and matching token relationships within the table sequences.

\begin{figure}[t]
    \centering
    \hspace{-0.2cm}\includegraphics[width=0.49\textwidth]{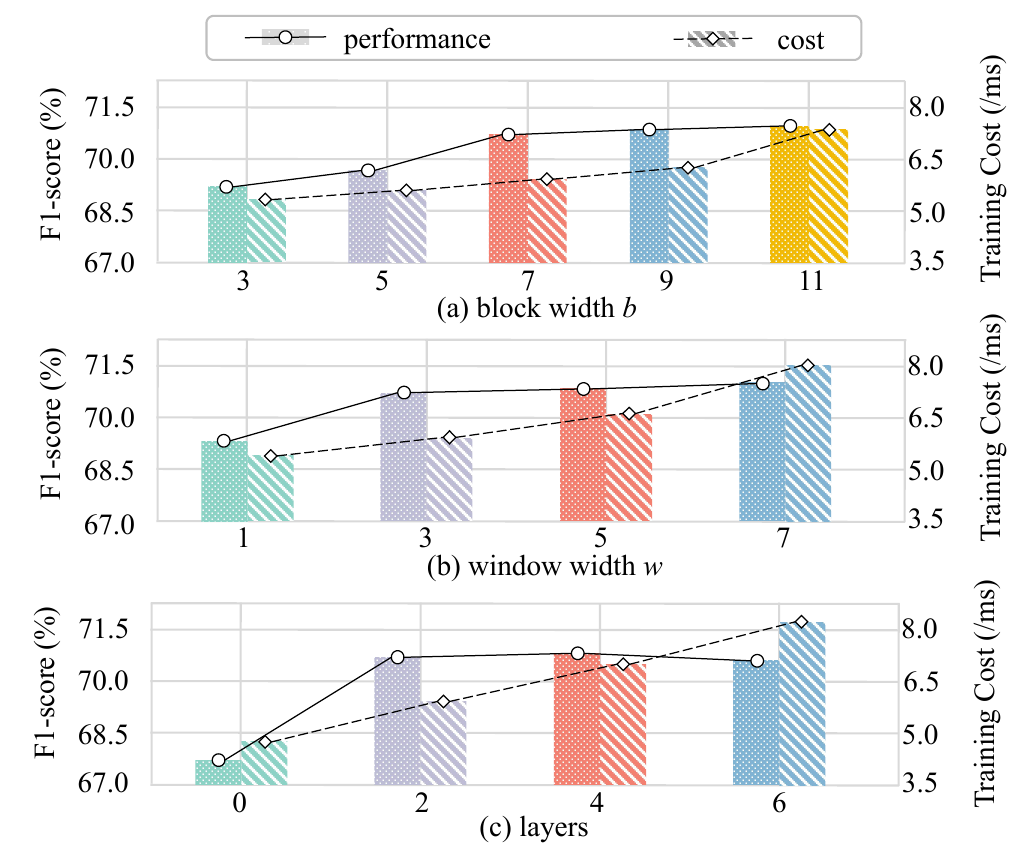}
    \vspace{-0.5cm}
    \caption{The sensitivity of different hyperparameters.}
    \label{fig:4}
\end{figure}
\subsection{Hyperparameter Analysis}
To investigate the impact of different hyperparameter settings in the T-T module on model performance, we conduct an additional experiment. 
The results, depicted in Figure \ref{fig:4}, demonstrate the effects when varying (a) the block width $b$, (b) the window width $w$, and (c) the T-T layer count while keeping other settings consistent with the main experiment.

In Figure \ref{fig:4} (a), as the block width b increases, the training time cost exhibits a nearly quadratic growth, consistent with the $O(w^2b^2n^2)$ complexity of the stripe attention mechanism as stated earlier. 
Additionally, when the block width exceeds 7, there is minimal improvement in the model's performance. 
We attribute this to the observation range ($b\times w=3 \times 7$) approaching saturation, effectively attending to all important tokens.
The findings drawn from Figure \ref{fig:4} (b) are consistent with Figure \ref{fig:4} (a) as the training time cost exhibits a quadratic increase with the window width $w$. 
Additionally, when the window width exceeds 3, there is minimal performance improvement.
In Figure \ref{fig:4} (c), as the number of T-T layers increases, the computational cost shows linear growth. 
When the number of layers exceeds 2, there is a slight decline in model performance, which we attribute to potential underfitting caused by an excessive number of parameters.

\subsection{Further Analysis on Loop-shift Strategy}
\begin{figure}[t]
    \centering
    \includegraphics[width=0.48\textwidth]{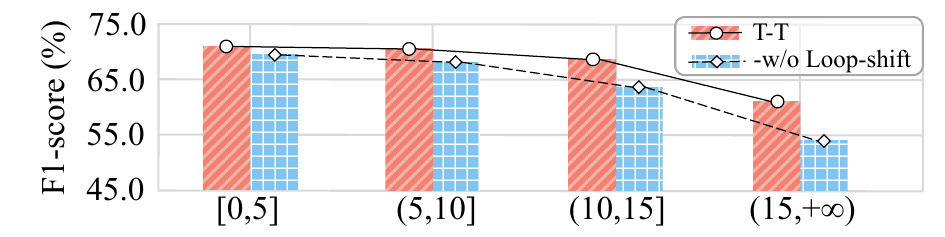}
    \vspace{-0.4cm}
    \caption{F1 scores for different aspect-opinion word distances on the test set. The sample counts in different distance intervals are 2442, 449, 107, and 52, respectively.}
    \label{fig:5}
\end{figure}

The loop-shift strategy is employed to address the challenge of \textbf{unfair local attention interaction}. We conducted an additional experiment to validate this strategy further.
As shown in Figure \ref{fig:5}, removing the loop-shift strategy results in a significant performance drop of the model on long-distance pairs.
This demonstrates the effectiveness of the loop-shift strategy, as removing it prevents tokens at the boundaries of attention windows from fully capturing valuable information, resulting in significant performance degradation.

\begin{figure}[t]
    \centering
    \includegraphics[width=0.49\textwidth]{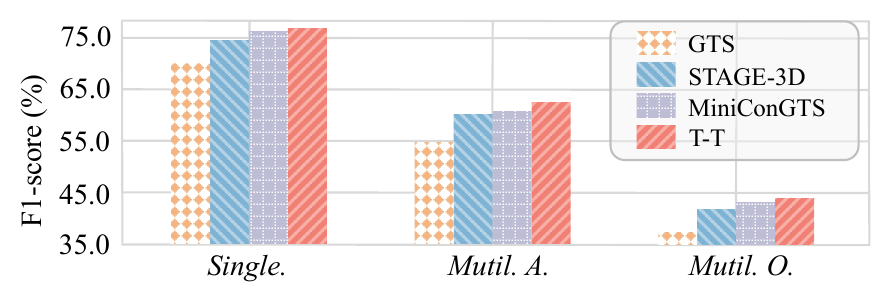}
    \vspace{-0.5cm}
    \caption{Performance of different word spans.  \textit{Single.} denotes triplets with single-word aspects and opinions. \textit{Multi. A.}/\textit{Multi. O.} denote triplets with multiple-word aspects/opinions.}
    \label{fig:6}
\end{figure}

\subsection{Performance of Different Word Spans}
We also compare the performance of T-T with other table tagging methods across different word spans, including single-word (\textit{Single.}), multi-word aspect (\textit{Mutil. A.}), and multi-word opinion (\textit{Mutil. O.}).
The results are shown in Figure \ref{fig:6}. 
Our model outperforms previous methods in all settings, with the improvement being more pronounced in the multi-word setting. 
This highlights the ability of T-T to effectively capture word boundary information.

\section{Conclusion}\label{sec:conclusion}
In this paper, we propose a novel Table-Transformer (T-T) approach, which uses enhanced transformer layers as the relation encoding module for the table-based ASTE task.
To address the challenges posed by overly long table sequences and hard local attention interaction, we respectively propose our stripe attention mechanism and the loop-shift strategy.
The former reduces attention computation costs by focusing on local tokens within the 2D table space, while the latter facilitates interaction between attention windows through loop-shift operations.
Extensive experiments on four datasets demonstrate the effectiveness of our T-T over the best baselines.
In future work, we aim to adapt this method to a broader range of information extraction tasks, such as relation extraction and event extraction.


\section*{Acknowledgments}
This research is supported by the National Key R\&D Program of China (No. 2023YFC3303800), NSFC through grants 62322202 and 62441612, ``Pioneer and Leading Goose R\&D Program of Zhejiang" through grant 2025C02044, National Key Laboratory under grant 241-HF-D07-01, and Hebei Natural Science Foundation through grant F2024210008.

\bibliographystyle{named}
\bibliography{6.ref}

\end{document}